\documentclass{article} 
\usepackage{arxiv,times}

\usepackage{hyperref}
\usepackage{url}

\usepackage[utf8]{inputenc} 
\usepackage[T1]{fontenc}    
\usepackage{booktabs}       
\usepackage{amsfonts,amsmath,amssymb} 
\usepackage{xcolor}         
\usepackage{tikz}

\title{Semi-supervised learning objectives as\\log-likelihoods in a generative model of data curation}

\author{  Stoil Ganev\\
  Department of Computer Science\\
  University of Bristol, Bristol, UK\\
  \texttt{stoil.ganev@bristol.ac.uk}
  \And
  Laurence Aitchison\\
  Department of Computer Science\\
  University of Bristol, Bristol, UK\\
  \texttt{laurence.aitchison@bristol.ac.uk}
}

\newcommand{\bcket}[3]{\left#1 #3 \right#2}
\newcommand{\mbcket}[5]{\left#1 #4 \middle#2 #5 \right#3}
\renewcommand{\b}{\bcket{(}{)}}

\newcommand{\sqb}{\bcket{[}{]}}
\newcommand{\sqbc}{\mbcket{[}{\vert}{]}}
\newcommand{\cb}{\bcket{\{}{\}}}

\renewcommand{\P}[1][]{\operatorname{P}_{#1}\b}

\renewcommand{\ss}[1]{\{ #1 \}_{s=1}^S}

\newcommand{\tsum}{\textstyle{\sum}}

\newcommand{\E}[1][]{\mathbb{E}_{{#1}} \sqb}
\newcommand{\Ec}[1][]{\mathbb{E}_{{#1}} \sqbc}

\DeclareMathOperator*{\argmax}{arg\,max}

\newcommand{\ceq}{\mkern2mu{=}\mkern1mu}

\newcommand{\obs}{\textcolor{red}}

\iclrfinalcopy 
\begin{document}

\maketitle
\begin{abstract}
We currently do not have an understanding of semi-supervised learning (SSL) objectives such as pseudo-labelling and entropy minimization as log-likelihoods, which precludes the development of e.g.\ Bayesian SSL.
Here, we note that benchmark image datasets such as CIFAR-10 are carefully curated, and we formulate SSL objectives as a log-likelihood in a generative model of data curation that was initially developed to explain the cold-posterior effect (Aitchison 2020).
SSL methods, from entropy minimization and pseudo-labelling, to state-of-the-art techniques similar to FixMatch can be understood as lower-bounds on our principled log-likelihood.
We are thus able to give a proof-of-principle for Bayesian SSL on toy data. 
Finally, our theory suggests that SSL is effective in part due to the statistical patterns induced by data curation.
This provides an explanation of past results which show SSL performs better on clean datasets without any ``out of distribution'' examples.
Confirming these results we find that SSL gave much larger performance improvements on curated than on uncurated data, using matched curated and uncurated datasets based on Galaxy Zoo 2.\footnote{Our code: https://anonymous.4open.science/r/GZ\_SSL-B6CC; MIT Licensed}
\end{abstract}

\section{Introduction}

To build high-performing deep learning models for industrial and medical applications, it is necessary to train on large human-labelled datasets.
For instance, Imagenet \citep{deng2009imagenet}, a classic benchmark dataset for object recognition, contains over 1 million labelled examples.
Unfortunately, human labelling is often prohibitively expensive. 
In contrast obtaining unlabelled data is usually very straightforward.
For instance, unlabelled image data can be obtained in almost unlimited volumes from the internet.
Semi-supervised learning (SSL) attempts to leverage this unlabelled data to reduce the required number of human labels \citep{seeger2000learning,zhu2005semi,chapelle2006semi,zhu2009introduction,van2020survey}.
One family of SSL methods --- those based on low-density separation --- assume that decision boundaries lie in regions of low probability density, far from all labelled and unlabelled points.
To achieve this, pre deep learning (DL) low-density separation SSL methods such as entropy minimization and pseudo-labelling \citep{grandvalet2005semi,lee2013pseudo} use objectives that repel decision boundaries away from unlabelled points by encouraging the network to make more certain predictions on those points.
Entropy minimization (as the name suggests) minimizes the predictive entropy, whereas pseudo-labelling treats the currently most-probable label as a pseudo-label, and minimizes the cross entropy to that pseudo-label.
More modern work uses the notion of consistency regularisation, which augments the unlabelled data (e.g. using translations and rotations), then encourages the neural network to produce similar outputs for different augmentations of the same underlying image \citep{sajjadi2016regularization,xie2019unsupervised,berthelot2019mixmatch,sohn2020fixmatch}.
Further developments of this line of work have resulted in many variants/combinations of these algorithms, from directly encouraging the smoothness of the classifier outputs around unlabelled datapoints \citep{miyato2018virtual} to the ``FixMatch'' family of algorithms \citep{berthelot2019mixmatch,berthelot2019remixmatch,sohn2020fixmatch}, which combine pseudo-labelling and consistency regularisation by augmenting each image twice, and using one of the augmented images to provide a pseudo-label for the other augmentation.

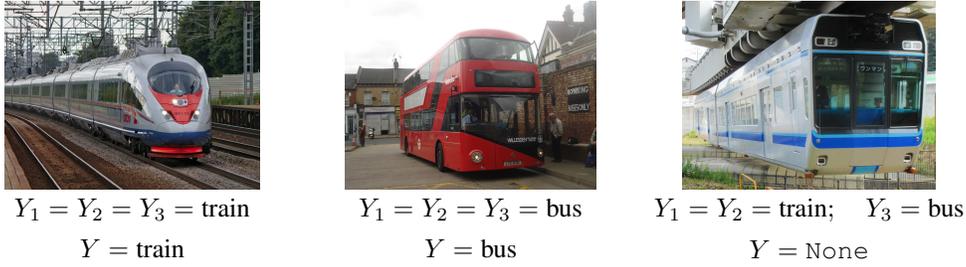
\begin{figure*}
  \centering
  \begin{tikzpicture}
    \node[inner sep=0pt] (train) at (0,0)
      {\includegraphics[height=2.5cm, trim={200 0 0 0},clip]{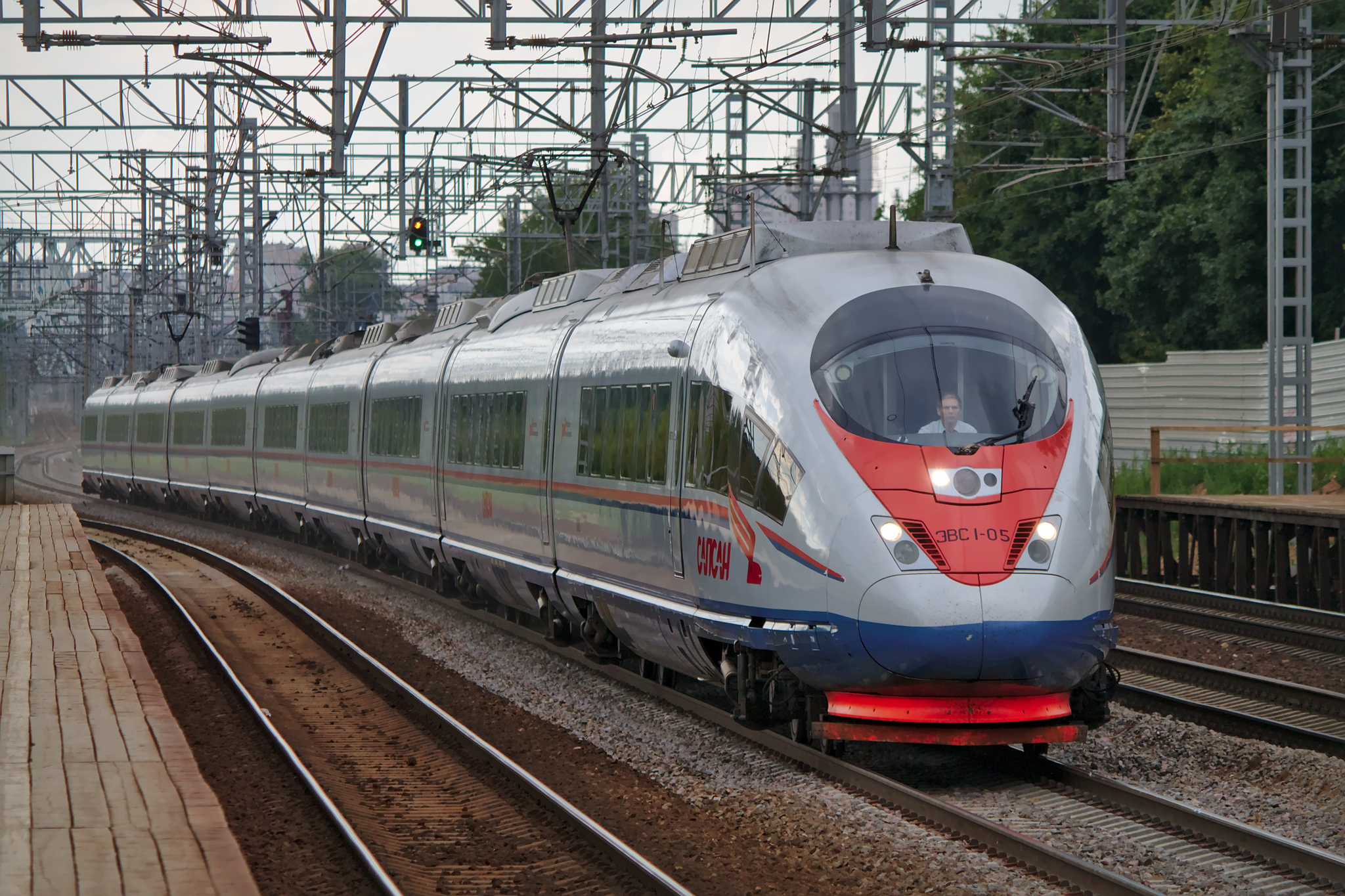}};
    \node[anchor=north] (lab) at (train.south) {$Y_1=Y_2=Y_3=\text{train}$};
    \node[anchor=north] (lab) at (lab.south) {$Y=\text{train}$};
    
    \node[inner sep=0pt] (bus) at (4.5cm,0)
      {\includegraphics[height=2.5cm]{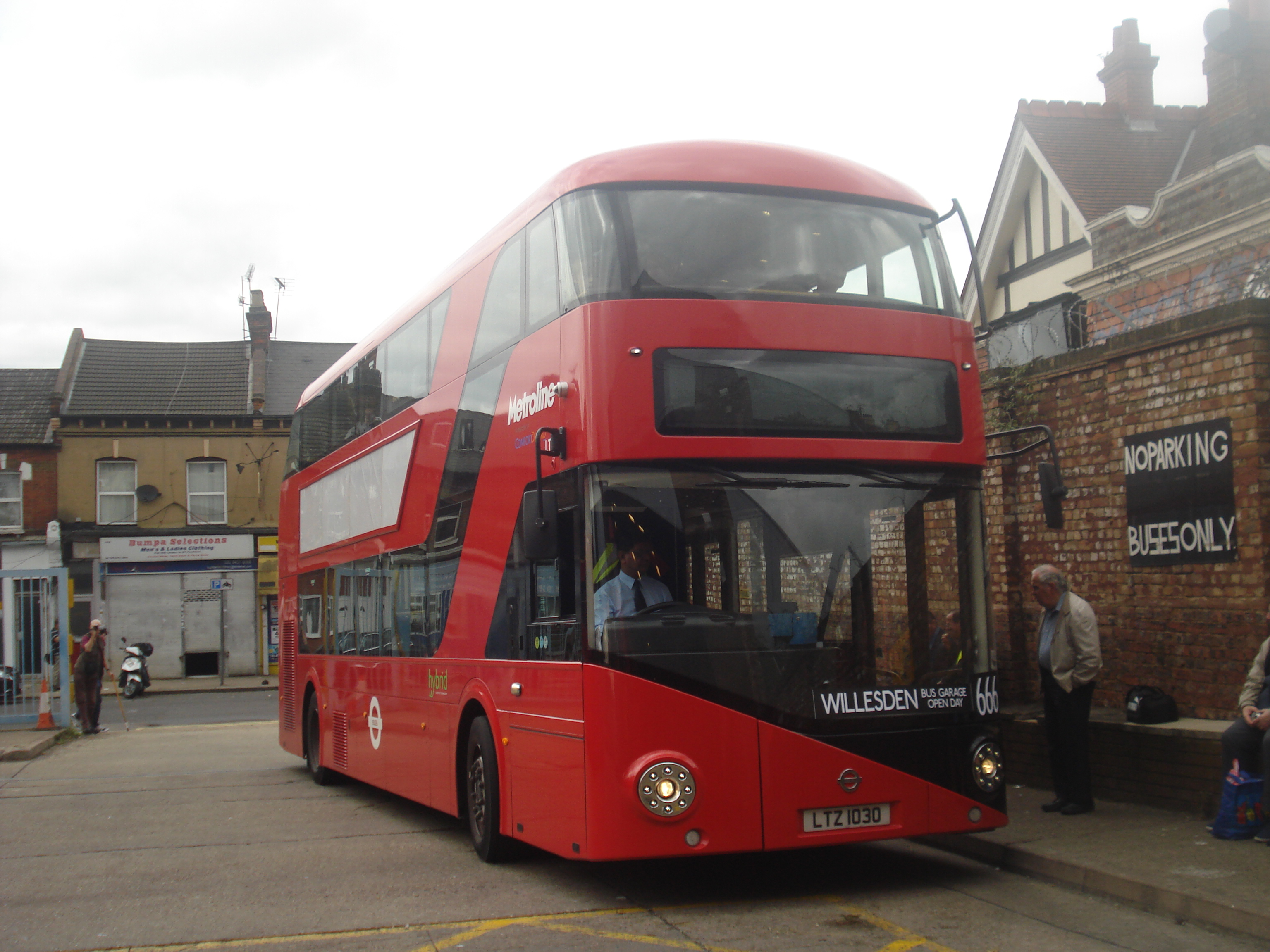}};
    \node[anchor=north] (lab) at (bus.south) {$Y_1=Y_2=Y_3=\text{bus}$};
    \node[anchor=north] (lab) at (lab.south) {$Y=\text{bus}$};
    \node[inner sep=0pt] (amb) at (9cm,0)
      {\includegraphics[height=2.5cm]{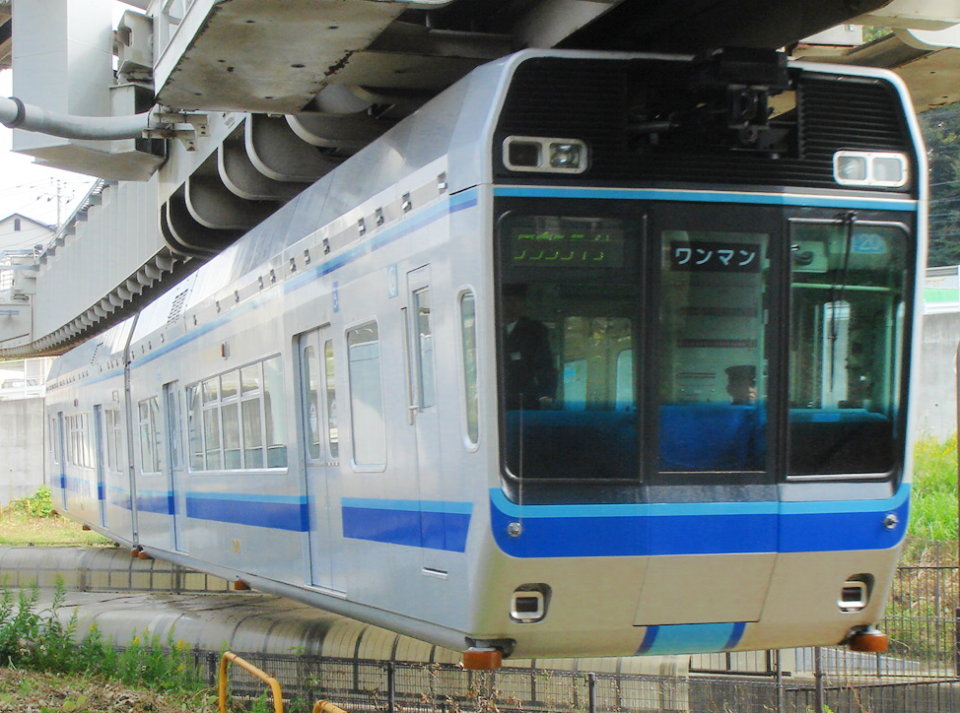}};
    \node[anchor=north] (lab) at (amb.south) {$Y_1=Y_2=\text{train};\quad Y_3=\text{bus}$};
    \node[anchor=north] (lab) at (lab.south) {$Y=\texttt{None}$};
  \end{tikzpicture}
  \caption{
    A depiction of the generative model of data curation, with $S=3$.
    annotators are instructed to classify images as trains or buses.
    The left-hand image is clearly a train, so the annotators agree and consensus is reached.
    The middle image is clearly a bus, so annotators agree and consensus is reached.
    The right-hand image, however, is ambiguous or even has an ill-defined class label.  So annotators disagree, consensus is not reached and the image is excluded from the dataset.
    \label{fig:schematic}
  }
\end{figure*}
\begin{figure*}
  \centering
  \begin{tikzpicture}
    \def\dx{1.5cm}
    \def\dy{0.7cm}
    \def\df{4cm}
    \def\da{-4cm}
    \def\db{-3cm}
    \node (A) at ({\da+\db-0.5*\dx}, 0) {\textbf{A}};
    \node (X) at ({\da+\db}, 0) {$\obs{X}$};
    \node (W) at ({\da+\db}, {-2*\dy}) {$\theta$};
    \node (Y) at ({\da+\db+\dx}, {-\dy}) {$\obs{Y}$};
    \draw[->] (X) -- (Y);
    \draw[->] (W) -- (Y);
    
    \node (B) at ({\db-0.5*\dx}, 0) {\textbf{B}};
    \node (X) at ({\db}, 0) {$\obs{X}$};
    \node (W) at ({\db}, {-2*\dy}) {$\theta$};
    
    \node (C) at ({-0.5*\dx}, 0) {\textbf{C}};
    \node (X) at ({0}, 0) {$\obs{X}$};
    \node (W) at ({0}, {-2*\dy}) {$\theta$};
    \node (Ys)at ({0+\dx}, {-\dy}) {$\ss{Y_s}$};
    \node (Y) at ({0+2*\dx}, {-1*\dy}) {$\obs{Y}$};
    
    \draw[->] (X) -- (Ys);
    \draw[->] (W) -- (Ys);
    \draw[->] (Ys) -- (Y);
    \draw[->] (Ys) -- (Y);
    
%
  \end{tikzpicture}
  \caption{
    Graphical models under consideration.  
    The observed variables are highlighted in red.
    \textbf{A} The generative model for standard supervised learning with no data curation.
    \textbf{B} The generative model for standard supervised learning, omitting the label.
    \textbf{C} The generative model with data curation, where the noconsensus images are observed.
    \citep[Adapted with permission from][]{aitchison2020tempering}.
    \label{fig:graphical_model}
  }
\end{figure*}
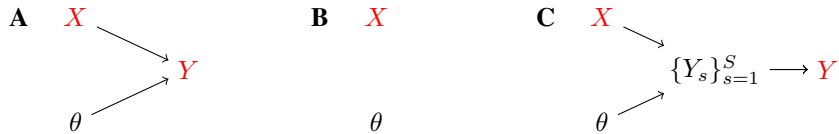
However, some of the biggest successes of deep learning, from supervised learning to many generative models, have been built on a principled statistical framework as maximum (marginal) likelihood inference (e.g.\ the cross-entropy objective in supervised learning can be understood as the log-likelihood for a Categorical-softmax model of the class-label \citealp{mackay2003information}).
Low-density separation SSL methods such as pseudo-labelling and entropy minimization are designed primarily to encourage the class-boundary to lie in low-density regions. 
Therefore they cannot be understood as log-likelihoods and cannot be combined with principled statistical methods such as Bayesian inference.

Here, we give a formal account of SSL methods based on low-density separation \citep{chapelle2006semi} as lower bounds on a principled log-likelihood.
In particular, we consider pseudo-labelling \citep{lee2013pseudo}, entropy minimization \citep{grandvalet2005semi}, and modern methods similar to FixMatch \citep{sohn2020fixmatch}.
This log-likelihood arises from a generative model of data curation that was initially developed to explain the cold-posterior effect \citep{aitchison2020tempering}.
Critically, this approach gives an explanation for previous findings that SSL is most effective when unlabelled data is obtained by throwing away labels from the carefully curated training set, and is less effective when unlabelled data is taken from uncurated images, especially those that do not depict one of the classes of interest \citep{oliver2018realistic,chen2020semi,guo2020safe}.
We confirmed the importance of data curation for SSL on toy data generated from a known model and on real data from Galaxy Zoo 2 \citep{willett2013galaxy}.

\section{Background}
Our work brings together many disparate areas.
Here, we give an introduction to a generative model of data curation \citep{aitchison2020tempering} initially developed to explain the cold posterior effect \citep{wenzel2020good}, pseudo-labelling and entropy minimization \citep{grandvalet2005semi,lee2013pseudo}, and the treatment of unlabelled points in the standard supervised learning setup.

\subsection{A generative model of data curation}
\label{sec:back:curation}
\begin{figure*}
  \centering
  \includegraphics[width=0.7\textwidth]{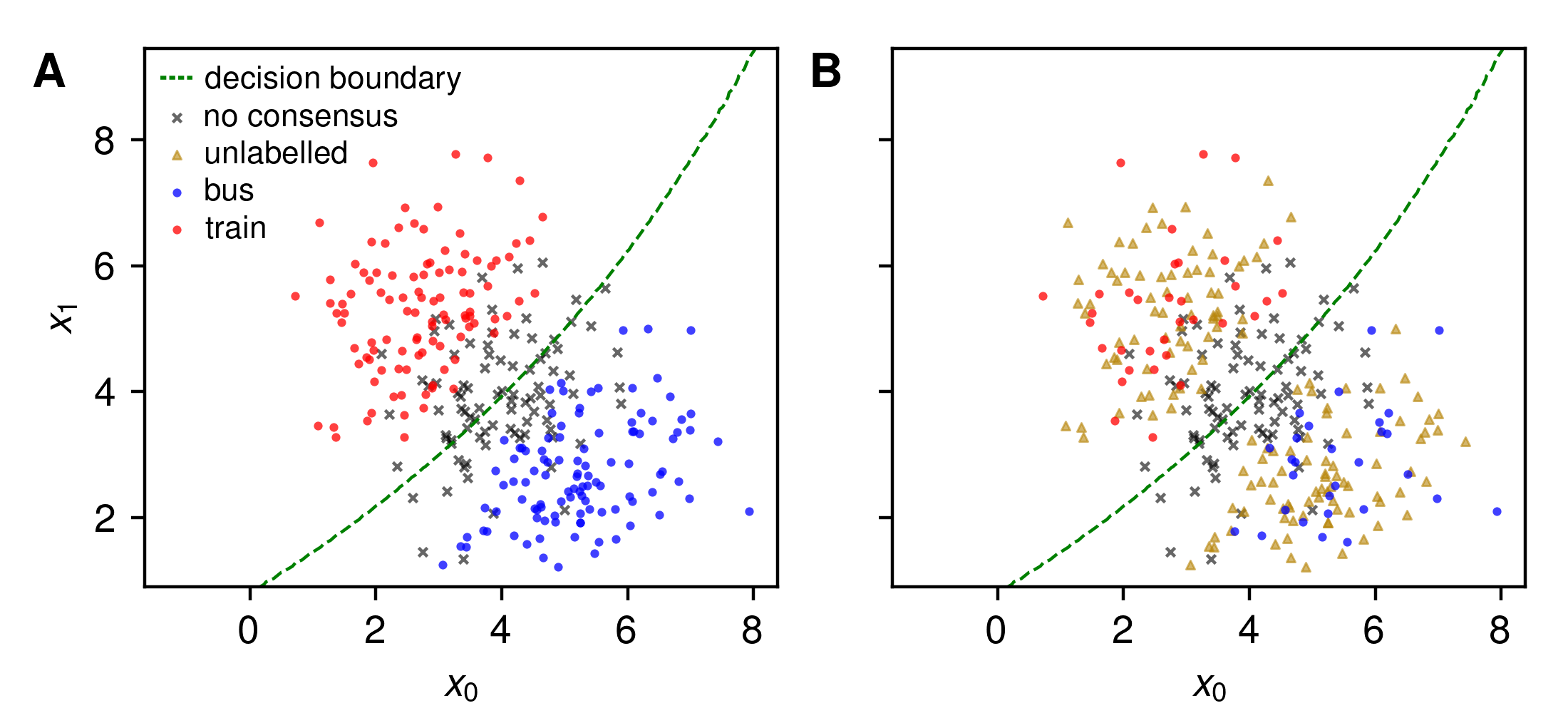}
  \caption{
    Our generative model of data curation applied to a simple 2D dataset.
    Data from each class was sampled from a different Gaussian, and the true decision boundary (green dashed line) was given by the posterior probability of class given $(x_0, x_1)$.
    \textbf{A} Datapoints far from the decision boundary are unambiguous, so annotators agree and consensus is reached (red and blue points).  Datapoints close to the decision boundary are ambiguous, so consensus is not reached (grey crosses).
    The consensus datapoints thus exhibit artificially induced low-density separation.
    \textbf{B} When using benchmark datasets such as CIFAR-10, the unlabelled points (yellow triangles) are selected from the consensus points (red or blue points) as the noconensus points are not available.  The unlabelled points therefore also exhibit artificially induced low-density separation.
    \label{fig:gen_model}
  }
\end{figure*}

To develop a model of data curation, remember that image datasets including CIFAR-10 and ImageNet are curated to ensure sure they only contain images whose class-labels are unambiguous.
For instance, in CIFAR-10, annotators were instructed that ``It's worse to include one that shouldn't be included than to exclude one.'', and \citet{krizhevsky2009learning} ``personally verified every label submitted by the annotators''.
In creating ImageNet, \citet{deng2009imagenet} made sure that a number of Amazon Mechanical Turk annotators agreed upon the class before including an image in the dataset.

Thus, these datasets have two odd properties.
First, consensus labels exist only for a subset of images, e.g. for a white-noise image, consensus cannot be reached and the image cannot be labelled.
Second, inclusion of an image in a dataset like CIFAR-10 is informative in and of itself, as it indicates that the image shows an unambiguous example of one of the ten classes.
To understand these odd properties of curated datasets, consider a simplified generative model of consensus-formation: draw a random image, $X$, from the distribution over images, $\P{X}$, and ask $S$ human annotators, indexed $s$, to give a label, $\ss{Y_s}$ (e.g. using Mechanical Turk). 
Importantly, every annotator is forced to label every image and if the image is ambiguous they should give a random label.
If all the annotators agree, $Y_1\ceq Y_2\ceq\dotsm\ceq Y_S$, they have consensus and the datapoint is included in the dataset. 
However, in the case of any disagreement, consensus is not reached and the datapoint is excluded (Fig.~\ref{fig:schematic}),
Concretely, the final label, $Y$ is $Y_1$ (which is the same as all the other labels) if consensus was reached and $\texttt{None}$ otherwise (Fig.~\ref{fig:graphical_model}C),
\begin{align}
  Y | \ss{Y_s} &= \begin{cases}
    Y_1 & \text{if } Y_1\ceq Y_2\ceq \dotsm \ceq Y_S\\
    \texttt{None} & \text{otherwise}
  \end{cases}
\end{align}
Taking $\mathcal{Y}$ to be the label set, we have $Y_s \in\mathcal{Y}$, and the final label, $Y$, could be any of the underlying labels in $\mathcal{Y}$, or \texttt{None} if consensus is not reached, so $Y \in \mathcal{Y} \cup \cb{\texttt{None}}$.
When consensus was reached, the likelihood is,
\begin{align}
  \P{Y\ceq y| X, \theta} &= \P{\ss{Y_s \ceq  y}| X, \theta} 
  = {\textstyle\prod}_{s=1}^S \P{Y_s\ceq y| X, \theta} 
  \label{eq:con_like}
  = \P{Y_s\ceq y| X, \theta}^S = p^S_y(X)
\end{align}
where we have assumed annotators are IID, and $p_y(X)=\P{Y_s\ceq y| X, \theta}$ is the single-annotator probability.
From here, it is possible to see how this model might be taken to give an account of tempering, as we have taken the underlying single-annotator likelihood, $p_y(X)$ to the power $S$ \citep[for further details see][]{aitchison2020tempering}.


\subsection{Low-density separation semi-supervised learning objectives}
The intuition behind low-density separation objectives for semi-supervised learning is that decision boundaries should be in low-density regions away from both labelled and unlabelled data.
As such, it is sensible to ``repel'' decision boundaries away from labelled and unlabelled datapoints and this can be achieved by making the classifier as certain as possible on those points.
This happens automatically for labelled points as the standard supervised objective encourages the classifier to be as certain as possible about the true class label.
But for unlabelled points we need a new objective that encourages certainty, and we focus on two approaches.
First, and perhaps most direct is entropy minimization \citep{grandvalet2005semi}
\begin{align}
  \mathcal{L}_\text{entropy}(X) &= \sum_{y \in \mathcal{Y}} p_y(X) \log p_y(X)
\end{align}
where, following the typical probabilistic approach, we write the negative entropy as an objective to be maximized.
Alternatively, we could use pseudo-labelling, which takes the current classification, $y^*$, to be the true label, and maximizes the log-probability of that label \citep{lee2013pseudo},
\begin{align}
  \mathcal{L}_\text{pseudo}(X) &= \log p_{y^*}(X) & y^* &= \argmax_{y\in\mathcal{Y}} \log p_{y}(X)
\end{align}
While these two approaches undeniably increase certainty and thereby repel the decision boundary, they are not formulated as a principled log-likelihood, which gives rise to at least three problems.
First, these methods cannot be combined with other principled statistical methods such as Bayesian inference.
Second, it is unclear how to combine these objectives with standard supervised objectives, except by taking a weighted sum and doing hyperparameter optimization over the weight.
Third, these objectives risk reinforcing any initial poor classifications and it is unclear whether this is desirable.

\subsection{In standard supervised learning, unlabelled points should be uninformative}
\begin{figure}
  \hspace{0.2in} \textbf{A} \hspace{3.25in} \textbf{B}
  
  \centering
  \includegraphics[height=1.8in]{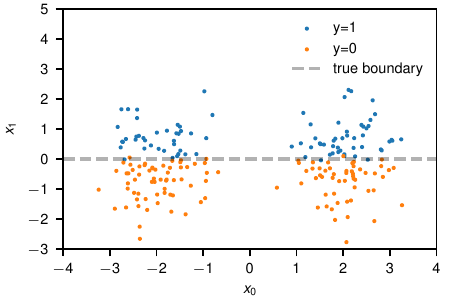}
  \hspace{0.7in}
  \includegraphics[height=1.8in]{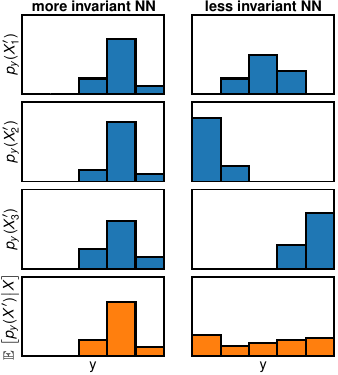}
  \caption{
    \textbf{A}. A toy dataset generated to illustrate the dangers of using the clustering of the input points to inform classification boundaries.
    The input features, $x_0$ and $x_1$ are plotted on the x and y-axes and the class is represented by colour. 
    \textbf{B}. A schematic diagram demonstrating the effect of our principled likelihood incorporating data-augmentation on the certainty of predictions for different degrees of invariance.
    More invariant NNs (left) give similar predictive distributions for different augmentations (blue), and hence a certain averaged predictive distribution (bottom; orange).
    Less invariant NNs (right) give different predictive distributions for different augmentations (blue), and hence highly uncertain averaged predictive distributions (bottom; orange).
    \label{fig:cluster_not_class}
  }
\end{figure}

It is important to note that under the standard supervised-learning generative model (Fig.~\ref{fig:graphical_model}A), unlabelled points should not give any information about the weights.
Omitting the label, $Y$, we obtain the graphical model in Fig.~\ref{fig:graphical_model}B. 
This model emphasises that the images, $X$, and the model parameters, $\theta$, are marginally independent, so we cannot obtain any information about $\theta$ from $X$ alone (Fig.~\ref{fig:graphical_model}B).
Formally, the posterior over $\theta$ conditioned on $X$ is equal to the prior,
\begin{align}
  \P{\theta| X} &= \frac{\P{\theta, X}}{\P{X}} = \frac{\tsum_{y\in\mathcal{Y}} \P{\theta, X, Y_s{=}y}}{\P{X}}
  = \frac{\P{\theta} \P{X} \sum_{y\in\mathcal{Y}} \P{Y_s{=}y| \theta, X}}{\P{X}} = \P{\theta}.
\end{align}
as $1 = \tsum_{y\in\mathcal{Y}} \P{Y_s{=}y| \theta, X}$.
To confirm this result is intuitively sensible, note that are many situations where encouraging the decision boundary to lie in low density regions would be very detrimental to performance.
Consider a classifier with two input features: $x_0$ and $x_1$ (Fig.~\ref{fig:cluster_not_class}A).
The class boundary lies in the high-density region crossing both clusters, so to obtain a reasonable result, the classifier should ignore the low-density region lying between the clusters.
However, strong low-density separation SSL terms in the objective may align the cluster boundaries with the class boundaries, leading the classifier to wrongly believe that one cluster is entirely one class and the other cluster is entirely the other class.
In contrast, supervised learning without SSL will ignore clustering and obtain a reasonable answer close to the grey dashed line.
Importantly, this is just an illustrative example to demonstrate that without further assumptions, the standard supervised approach of ignoring unlabelled data is sensible; semi-supervised learning without loss of performance in such settings has been studied and is known as Safe SSL \citep{li2014towards,krijthe2014implicitly,kawakita2014safe,loog2015contrastive,krijthe2016pessimistic}.

\section{Methods}
SSL methods are usually applied to benchmark datasets such as CIFAR-10 or ImageNet.
These datasets were first carefully curated during the labelling process: (Fig.~\ref{fig:gen_model}A), implying that ambiguous images close to the decision boundary were excluded.
Critically, unlabelled points for these benchmark datasets are obtained by taking labelled points (which have reached consensus) and throwing away their labels (Fig.~\ref{fig:gen_model}B).
The likelihood for consensus ($Y{\neq}\texttt{None}$) is
\begin{align}
  \label{eq:exact_consensus}
  \P{Y{\neq}\texttt{None}| X, \theta} &= \tsum_{y \in \mathcal{Y}} \; p_y^S(X).
\end{align}
This probability is close to $1$ (for $S>1$) if the underlying distribution, $p_y^S(X)$ puts most of its mass onto one class, and the probability is smaller if the mass is spread out over classes.
As such, the likelihood ``repels'' decision boundaries away from unlabelled points, which is the common intuition behind low-density separation SSL methods, and which should be beneficial if class boundaries indeed lie in regions of low probability density away from both labelled and unlabelled points.

If noconsensus images are observed (Fig.~\ref{fig:graphical_model}C), we can include a likelihood term for those images,
\begin{align}
  \P{Y\ceq\texttt{None}| X, \theta} &= 1-\P{Y\neq\texttt{None}| X, \theta}
  \label{eq:noconsensus}
  = 1 - \tsum_{y \in \mathcal{Y}} \; p_y^S(X).
\end{align}
If noconsensus images are not observed, we could in principle integrate over the underlying distribution over images, $\P{X\ceq x}$.
However, we do not even have samples from the underlying distributions over images (and if we did, we would have the noconsensus images so we could use Eq.~\ref{eq:noconsensus}).
As such this term is usually omitted \citep[e.g.][]{aitchison2020tempering}, but the use of out-of-distribution (OOD) datasets as surrogate noconsensus points is an important direction for future work. 

\subsection{Entropy minimization and pseudo-labels are lower bounds on our principled log-likelihood}
To prove that entropy minimization forms a lower-bound on our log-likelihood (Eq.~\ref{eq:exact_consensus}), we begin by writing the log-likelihood of consensus in terms of an expectation over labels, $y$,
\begin{align}
  \log \P{Y{\neq}\texttt{None}| X, \theta} &= \log \sum_{y\in\mathcal{Y}} p_y(X) p_y^{S-1}(X) = \log \E[p_y(X)]{p_y^{S-1}(X)}.
\end{align}
Applying Jensen's inequality, the negative entropy gives a lower-bound on our log-likelihood,
\begin{align}
  \nonumber
  \log \P{Y{\neq}\texttt{None}| X, \theta} &\geq \E[p_y(X)]{\log p_y^{S{-}1}(X)}\\
  &= (S{-}1) \tsum_{y\in\mathcal{Y}} \; p_y(X) \log p_y(X)
  = (S{-}1) \mathcal{L}_\text{entropy}(X)
\end{align}
Pseudo-labelling forms an alternative lower bound on the log-likelihood which is obtained by noting that all $p_y^S(X)$ are positive, so selecting any subset of terms in the sum gives a lower bound,
\begin{align}
  \log \P{Y{\neq}\texttt{None}| X, \theta} &= \log \tsum_{y\in\mathcal{Y}} \; p_y^S(X) 
  \geq \log p_{y^*}^S(X)
  = S \log p_{y^*}(X) 
  = S \mathcal{L}_\text{pseudo}(X).
\end{align}
The final equality holds if we choose $y^*$ to be the highest probability class, but the bound is valid for any choice of $y^*$.
As such, entropy minimization and pseudo-labelling optimize different lower-bounds on our principled log-likelihood, $\log \P{Y{\neq}\texttt{None}| X, \theta}$, which gives a potential explanation for the effectiveness of pseudo-labelling and entropy minimization.

\subsection{Data augmentation priors and FixMatch family methods}
FixMatch family methods combine data augmentation and pseudo-labelling.
To understand FixMatch as a bound on a principled log-likelihood, we therefore need a principled account of data augmentation as a likelihood.
Inspired by \citet{wenzel2020good} (their Appendix K), we consider a distribution, $\P{X'| X}$, over augmented images, $X'$, given the underlying unaugmented image, $X$.
We choose the single-annotator predictive distribution as the average over predictive distributions for many different augmented images,
\begin{align}
  \label{eq:data_aug:exact}
  \P{Y_s\ceq y| X, \theta} &= \Ec{p_y(X')}{X}
\end{align}
where $p_y(X')$ is the predictive probabilities resulting from applying the neural network to the augmented image. 
This is a sensible prior because we expect the neural network to be invariant under data-augmentation, and if the predictions are approximately invariant, then averaging the predictive distributions has little impact (Fig.~\ref{fig:cluster_not_class}B left).
However, if the predictions do vary dramatically with different data augmentations then we should not trust the network's classifications (i.e.\ we should have an uncertain predictive distribution), and averaging over very different predictive distributions for different augmentations indeed gives rise to broader, more uncertain predictions (Fig.~\ref{fig:cluster_not_class}B right).

\begin{figure*}
  \centering
  \includegraphics{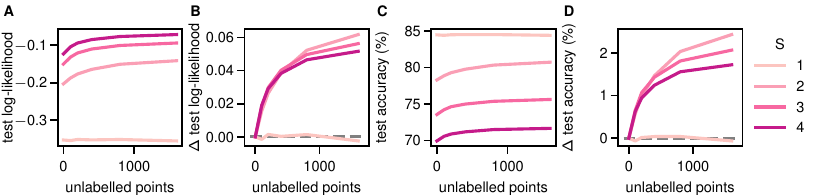}
  \caption{
    Test log-likelihood and accuracy for Langevin sampling for Bayesian SSL on toy datasets sampled from the model as a function of the number of unlabelled points.
    \label{fig:bayes} 
  }
\end{figure*}

To obtain a tractable objective in the supervised setting, we use a multi-sample version of Jensen's inequality, with $K$ augmented images denoted $X'_k$,
\begin{align}
  \label{eq:data_aug:multi_sample}
  \log\P{Y_s\ceq y| X, \theta} &\geq \Ec{\log \tfrac{1}{K} \textstyle{\sum}_k  p_y(X'_k)}{X}.
\end{align}
Combining this single-annotator probability with our generative model of curation, we obtain, 
\begin{align}
  \nonumber
  \log\P{Y\ceq y| X, \theta} &= S \log\P{Y_s\ceq y| X, \theta}\\
  &= S \log \Ec{p_y(X')}{X}
  \geq S \; \Ec{\log \tfrac{1}{K} \textstyle{\sum}_k  p_y(X'_k)}{X},
\end{align}
The resulting objective for unlabelled points is, 
\begin{align}
  \nonumber
  \log\P{Y{\neq}\texttt{None}| X, \theta} &= \log \tsum_{y\in\mathcal{Y}}\P{Y\ceq y| X, \theta}\\
  \label{eq:unlab-ssl-exact}
  &= \log \tsum_{y\in\mathcal{Y}}\Ec{p_y(X')}{X}^S
  \approx \log \tsum_{y\in\mathcal{Y}} \b{\tfrac{1}{K} \textstyle{\sum}_k  p_y(X'_k)}^S,
\end{align}
where we approximate the expectation with $K$ different samples of $X'$, denoted $X'_k$.
Unfortunately, this approach does not immediately form a bound on the log-likelihood due to the convex nonlinearity in taking the power of $S$.
Nonetheless, one key problem with approximating machine learning losses is that the optimizer learns to exploit approximation errors to find a pathological solution that makes the objective unboundedly large.
We appear to be safe from that pathology here, as we are simply forming predictions by averaging over $K$ augmentations of the underlying image.
Nonetheless, to form a lower bound, we can follow FixMatch family algorithms by psuedo-labelling, i.e.\ by taking only one term in the sum for class $y^*$.
%
FixMatch chooses $y^*$ by using the highest-probability class for a weakly-augmented image.
An alternative approach is to choose the $y^*$ giving the tightest bound, i.e.\ $\argmax_y \tfrac{1}{K} \textstyle{\sum}_k  p_y(X'_k)$.
In either case,
\begin{align}
  \label{eq:unlab-ssl-psuedo} 
  \log\P{Y{\neq}\texttt{None}| X, \theta} &\geq \log \Ec{p_{y^*}(X')}{X}^S
  \geq S\; \Ec{\log \tfrac{1}{K} \textstyle{\sum}_k  p_{y^*}(X'_k)}{X},
\end{align}
If $K=1$ and $y^*$ is chosen using a separate ``weak'' augmentation, then this is exactly equal to the FixMatch objective for unlabelled points.

Note that both of these objectives (Eq.~\ref{eq:unlab-ssl-exact}~and~\ref{eq:unlab-ssl-psuedo}) promote reduced predictive uncertainty.
Importantly, this does not just increase confidence in the single-augmentation predictive distributions, $p_y(X'_k)$, but also increases alignment between the predictive distributions for different augmentations (Fig.~\ref{fig:cluster_not_class}B).
In particular, if the single-augmentation predictives are all highly confident, but place that high-confidence on different classes, then the multi-augmentation predictive formed by averaging will have low-confidence (Fig.~\ref{fig:cluster_not_class}B right).
The only way for the multi-augmentation predictive to have high confidence is if the underlying single-augmentation predictive distributions have high confidence in the same class (Fig.~\ref{fig:cluster_not_class}B left), which encourages the underlying network to become more invariant.
This makes sense: if data-augmentation changes the class predicted by the neural network, then any predictions \textit{should} be low confidence.
And it implies that combining principled data augmentation with a generative model of data curation automatically gives rise to an objective encouraging invariance.

\section{Results}

We begin by giving a proof-of-principle for Bayesian SSL on a toy dataset generated from a known model.
Next, we tested our theoretical results (rather than trying to achieve SOTA performance) on real-world datasets.
In particular, our theory gives one explanation for why SSL is typically more effective when unlabelled data is taken from the original, curated training set. 
To confirm these results, we used Galaxy Zoo 2 as this was a real-world dataset which allowed us to generate matched curated and uncurated datasets. 

\subsection{Bayesian SSL on a generated dataset}

Our formulation of SSL as a likelihood implies that it should be possible to take entirely novel approaches, such as using low-density separation SSL in a Bayesian neural network (BNN).

We considered a toy dataset generated from a ``true'' neural network model with one hidden layer and 30 hidden units, 5 dimensional inputs and 2 output classes.
We generated inputs IID from a Gaussian, then passed them through the ``true'' neural network, then sampled multiple class-labels corresponding to different annotators.
If all the simulated annotators agreed, consensus was reached and if any simulated annotators disagreed, consensus was not reached.
We used 100 labelled datapoints, though not all of them will have reached consensus, and we used up to 1600 unlabelled points, though again not all of them will have reached consensus. 
Note that as the consensus/noconsensus status of a point arises from the generative model, we cannot independently specify the number of consensus/noconsensus points.
We used Eq.~\eqref{eq:con_like} as the likelihood for labelled points, Eq.~\eqref{eq:exact_consensus} as the likelihood for unlabelled points and Eq.~\eqref{eq:noconsensus} as the likelihood for noconsensus points.
We sampled (and trained networks on) 500 datasets in parallel.
We trained using Langevin dynamics with all data simultaneously (no minibatching) with no momentum and no rejection.

For a generative model with $S=1$, consensus is always reached and the problem is equivalent to standard supervised learning.
As such, we found no benefits from including unlabelled points for $S=1$.
In contrast, for any setting of $S>1$ we found that increasing the number of unlabelled points improved the test log-likelihood (Fig.~\ref{fig:bayes}AB) and the test accuracy (Fig.~\ref{fig:bayes}CD).

\subsection{Galaxy Zoo 2}
\begin{figure*}
  \centering
  \includegraphics{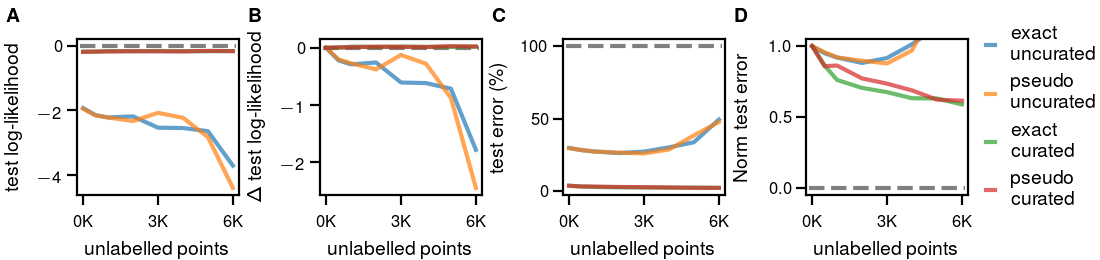}
  \caption{
    Test log-likelihood and error for curated and uncurated GZ2 datasets as a function of the number of unlabelled points.
    \label{fig:gz_sgd} 
  }
\end{figure*}

Our data-curation based theory predicts that SSL should be much more effective on curated than uncurated data.
To test this prediction on real-world data, we turned to Galaxy Zoo 2\footnote{https://data.galaxyzoo.org; www.sdss.org/collaboration/image-use-policy/} (GZ2) \citep{willett2013galaxy} which uses images from the Sloan Digital Sky Survey.
This dataset is particularly useful for us as it has received only very minimal filtering based on criteria such as object brightness and spatial extent.
We defined 9 labels by truncating the complex decision tree followed by the annotators \citep[for further details see][]{aitchison2020tempering}.
Further, as each GZ2 image has received $\sim 50$ labels, we can define a consensus coefficient by taking the fraction of annotators that agreed upon the highest probability class.
We can then define a curated dataset by taking the images with consensus coefficient above some threshold within each class.
Note that we needed to select images on a per-class basis, because annotators tend to be more confident on some classes than others, so taking the highest consensus coefficients overall would dramatically change the class balance.
In particular, we used the top $8.2\%$ of images, which gave a full curated dataset of just over 20,000 images.
Of those, we randomly selected 2000 as labelled examples, 10000 as test examples, and 0 -- 6000 as unlabelled examples.
The images were preprocessed by center-cropping to $212\times212$ and then scaled to $32 \times 32$.
We applied a FixMatch-inspired semi-supervised learning algorithm, with a standard supervised objective, with unlabelled objective given by Eq.~\eqref{eq:unlab-ssl-exact} with $K=2$. 
Data augmentation was given by vertical and horizontal flips, rotations from $-180^{\circ}$ to $180^\circ$, translations by up to 40\% on both axes and scaling from 20\% to 180\%.
Note that as we were trying to mirror the standard SSL setup, we did not include noconsensus points in the objective.
We trained a ResNet18 with our maximum likelihood objective using SGD with a batch size of 500, a learning rate of 0.01 and 1500 epochs.
We used an internal cluster of nVidia 1080 and 2080 GPUs, and the experiments took roughly 300 GPU hours.

We found that the test-log-likelihood for curated data improved slightly as more unlabelled points were included,  whereas the test-log-likelihood for uncurated dramatically declined as unlabelled points were added (Fig.~\ref{fig:gz_sgd}AB).
We saw strong improvements in test accuracy with the number of unlabelled points for curated datasets (Fig.~\ref{fig:gz_sgd}CD).
Note that in Fig.~\ref{fig:gz_sgd}C the error rate for curated datasets is already very small, so to see any effect we needed to plot the test error, normalized to the initial test error (Fig.~\ref{fig:gz_sgd}D).
For uncurated data, the inclusion of large numbers of unlabelled points dramatically worsened performance, though the inclusion of a small number of unlabelled points gave very small performance improvements (Fig.~\ref{fig:gz_sgd}CD).
Thus, this experiment is consistent with the idea that the effectiveness of SSL arises at least in part from curation of the underlying dataset.

\section{Related work}
There are at least three main approaches to semi-supervised learning \citep{seeger2000learning,zhu2005semi,chapelle2006semi,zhu2009introduction}.
First there is low-density separation, where we assume that the class boundary lies in a region of low probability density away from both labelled and unlabelled points.
This approach dates back at least to transductive support vector machines (SVMs) where the model is to be tested on a finite number of known test locations \citep{vapnik1998statistical,chapelle1999transductive}.
Those known test locations are treated as unlabelled points, and we find the decision boundary that perfectly classifies the limited number of labelled points, while at the same time being as far as possible from labelled and unlabelled data.
Alternative approaches include pseudo-labelling and entropy minimization \citep{grandvalet2005semi,lee2013pseudo}.
Second, there are graph-based methods such as \citep{zhu2002learning} which are very different from the methods considered here.
Third, there are approaches that use unlabelled points to build a generative model of the \textit{inputs} and leverage that model to improve classification \citep[e.g.][]{kingma2014semi,odena2016semi,gordon2017bayesian}.
This approach was originally explored in a considerable body of classical work \citep[e.g.][]{mclachlan1975iterative,castelli1995exponential,druck2007semi} for a review, see \citet{seeger2000learning} and references therein.
These approaches are fundamentally different from the SSL approaches considered here, as they require a generative model of inputs, while low-density separation methods do not.
Generative modelling can be problematic as training a generative model can be more involved than training a discriminative model and because the even when the model can produce excellent samples, the high-level representation may be ``entangled'' \citep{higgins2016beta} in which case it may not offer benefits for classification.

\section{Discussion}
\label{sec:discussion}

Our theory provides a theoretical understanding of past results showing that SSL is more effective when unlabelled data is drawn from the original, curated training set \citep{oliver2018realistic,chen2020semi,guo2020safe}.
In the extreme, our theory might be taken to imply that if data has not been curated, then SSL cannot work, and therefore that low-density separation SSL methods will not be effective in messy, uncurated real-world datasets.
However, this is not the complete picture.
Low-density separation SSL methods, including our log-likelihood, fundamentally exploit class-boundaries lying in low-density regions. 
As such, low-density separation could equally come from the real underlying data or could be artificially induced by data curation (Fig.~\ref{fig:gen_model}).
None of these methods are able to distinguish between these different underlying sources of low-density separation and as such any of them may work on uncurated data where the underlying distribution displays low-density separation.
However, the possibility for curation to artificially induce low-density separation does imply that we should be cautious about overinterpreting spectacular results obtained on very carefully curated benchmark datasets such as CIFAR-10.

Surprisingly, the generative model of data curation used here also explains the cold-posterior effect in Bayesian neural networks \citep{wenzel2020good,aitchison2020tempering}, revealing a profound and previously unsuspected connection.

\section{Conclusion}
We showed that low-density separation SSL objectives can be understood as a lower-bound on a log-probability which arises from a principled generative model of data curation.
This gives a theoretical understanding of recent results showing that SSL is more effective on curated data, which we confirmed
by developing a Bayesian SSL model applied to toy data, using GZ2, which allowed us to consider a completely uncurated dataset.

This work may have a social impact through the future development of enhanced SSL methods that reduce the cost of collecting large labelled training datasets

\bibliographystyle{iclr2022_conference}
\bibliography{refs}

\end{document}